\documentclass[letterpaper, 10pt, conference]{ieeeconf}

\IEEEoverridecommandlockouts
\usepackage{cite}
\usepackage{amsmath,amssymb,amsfonts}
\usepackage{algorithmic}
\usepackage{graphicx}
\usepackage{textcomp}
\usepackage{xcolor}
\usepackage{booktabs}
\usepackage{multirow}
\usepackage[hidelinks]{hyperref}
\usepackage{placeins}
\usepackage{paralist}

\begin{document}

\title{Mining Cross-Modal Perception Gaps in Autonomous Driving: When Camera and LiDAR Disagree}

\author{Hongsen Cao, Mona Jaber, Shanxin Yuan and Ahmed M. Abdelmoniem%
\thanks{The authors are with the School of Electronic Engineering and Computer Science, Queen Mary University of London, London, United Kingdom.
        {\tt\small \{hong.cao, m.jaber, shanxin.yuan, ahmed.sayed\}@qmul.ac.uk}}%
}

\maketitle

\begin{abstract}
Multi-modal fusion for 3D object detection assumes that cameras and LiDAR are broadly complementary. In practice, how much each modality is trusted at a given location is decided by learned weights that are neither characterized nor auditable. We label every ground-truth object by which modalities detected it, then mine the resulting 0.4M-object database across nuScenes, KITTI, and Waymo. Because the patterns shift with distance from the ego vehicle, we read the mined model through Spatially Stratified Feature Attribution (SSFA), which conditions Shapley attribution on distance zones and exposes structure that global attribution averages away.
Three assumptions built into the current fusion design do not hold. The modality-preference boundary tracks the object itself: a traffic cone is nearly six times as likely as a truck to be seen by the camera alone (16.8\% vs.\ 2.9\%), regardless of range. Camera-only detections \emph{peak} at 10--30\,m, inside the band usually ceded to LiDAR, and fall away on either side. The drivers of failure are not additive either: the distance$\times$density interaction alone outweighs the entire main effect of viewing direction. We further find that modality dominance is directionally non-stationary, with objects behind the vehicle being markedly more LiDAR-dominant than those in front.
We compile the mined structure into a Modality Reliability Map and inject it into four pretrained fusion detectors through a lightweight attention layer. This improves performance by up to 1.3 mean Average Precision (mAP) while fine-tuning under 4\% of their parameters in 2--3 GPU-hours. Our work lifts reliability out of the weights and treats it as a quantity that can be examined before deployment. That is what a safety measure for multi-modal perception must rest on.
\end{abstract}


\section{Introduction}

Cameras and LiDAR are the most critical sensor pair in autonomous driving. The camera supplies texture and semantics, and the LiDAR provides precise depth. Built on this complementarity assumption, camera--LiDAR fusion has become the dominant paradigm for 3D object detection (e.g., BEVFusion~\cite{bevfusion}, TransFusion~\cite{transfusion}, GAFusion~\cite{gafusion}). Yet fusion is not always effective. Recent work~\cite{ecfusion,fdsnet} treats modality disagreement as a signal for the architecture to handle, without first characterizing it. Adaptive fusion therefore lacks a principled basis for knowing when each modality fails.

This raises three questions that current fusion design leaves unanswered:
\begin{inparaenum}
\item[(1)] Under what conditions is a camera more reliable than LiDAR, and vice versa?
\item[(2)] Do these differences follow predictable spatial and attribute-dependent patterns, and where do the boundaries lie?
\item[(3)] Can the resulting structure be turned into a usable signal for fusion?
\end{inparaenum}
They are safety-critical. Over-trusting the weaker modality where it fails introduces false positives or missed objects exactly when perception matters most. A fusion detector resolves that trade-off inside learned weights, which cannot be queried before deployment. What the field works with instead are shared assumptions. LiDAR is held to be more reliable at short range and to degrade faster at long range~\cite{bevfusion,robustbenchfusion,dong2023corruption}, and the two modalities to compensate for each other~\cite{relifusion}. Both hold as broad tendencies, but a safety-critical system cannot act on a tendency. It needs the conditions under which each holds, and those are missing.

Two routes lead to an answer. One can open the fusion model and trace how it arrives at its weights, but the explanation is entangled with the architecture and does not transfer. The other leaves the models closed: run a camera-only and a LiDAR-only detector over the same scenes and record which succeeded object by object. Disagreement becomes a label on an object, and asking when a modality fails becomes asking which attributes accompany that failure. We take this route.

The Cross-Modal Perception Gap (CMPG) framework follows that route. It labels each object with the modalities that found it, then mines the resulting database to identify the attribute combinations underlying disagreement. The result is a prior that a fusion detector can use. One difficulty runs through all three stages. The structure is not fixed across space. Factors driving disagreement change with distance from the ego vehicle, a property we call \emph{spatial non-stationarity}. Heterogeneous attributes, intra-scene correlations, and class imbalance compound it. We validate on nuScenes~\cite{nuscenes}, KITTI~\cite{kitti}, and Waymo~\cite{waymo}.

The main contributions are:
\begin{itemize}
    \item We deliver an \emph{auditable} fusion prior. The Modality Reliability Map assigns every BEV cell a camera reliability and a LiDAR reliability. Prior-Guided Spatial Attention injects that field into a pretrained detector, fine-tuning under 4\% of the host's parameters in 2--3 GPU-hours. Across four state-of-the-art fusion detectors, the layer adds up to 1.3 mAP without altering the backbones, and the field it injects into is open to inspection.
    \item Three findings contradict assumptions embedded in current fusion design. \emph{First}, physical cross-section, not range, decides which sensor sees a target. \emph{Second}, the near field is not LiDAR's uncontested domain, since camera-only detections peak at 10--30\,m, where LiDAR returns thin out while image resolution still holds. \emph{Third}, the failure drivers are non-additive, so no single-variable rule can capture them. The distance$\times$density interaction alone exceeds the main effect of viewing direction. Alongside these three, we find a front-to-rear directional non-stationarity that distance-only analysis cannot see.
    \item The prior rests on evidence at scale, roughly 0.4M labelled objects across three sensor suites. We read them through SSFA, which exposes how the dominant failure factor reorders across distance zones. We release the labels and labeling scripts as CMPG-Bench (repository link withheld for anonymous review).
\end{itemize}

\begin{figure*}[!t]
\centering
\includegraphics[width=\textwidth]{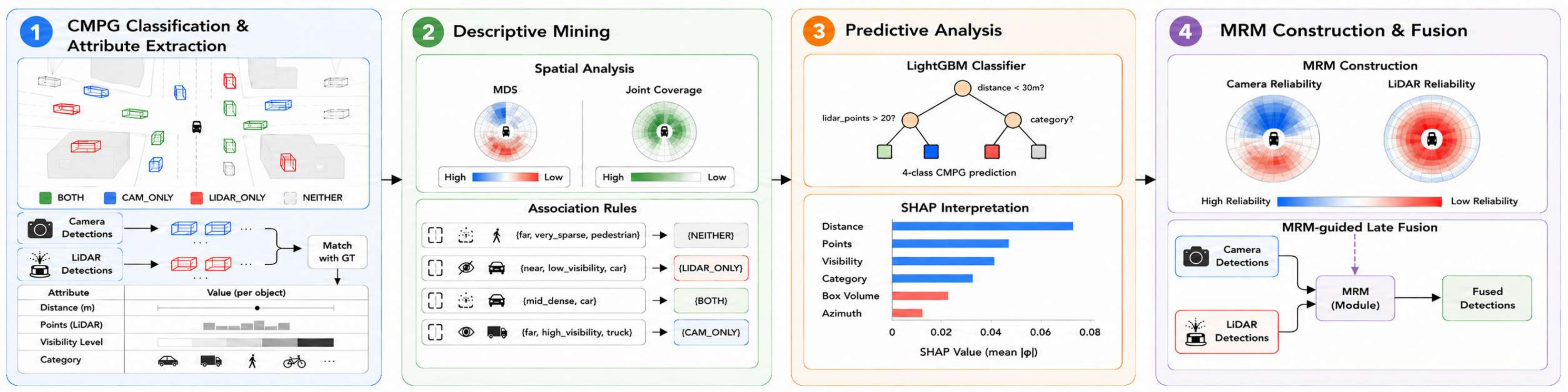}
\caption{The CMPG pipeline. Stages 1--3 discover the structure through classification, descriptive mining in bird's-eye-view (BEV), and attribution; stage 4 compiles it into a reliability map and injects it into a fusion detector.}
\label{fig:pipeline}
\end{figure*}

\section{Related Work}

\subsection{Adaptive Camera--LiDAR Fusion}
\label{sec:fusion3d}

Camera--LiDAR fusion has shifted from fixed-weight toward adaptive schemes. Single-modality detectors have meanwhile improved until their ranges overlap (camera: FCOS3D~\cite{fcos3d}, BEVFormer~\cite{bevformer}, StreamPETR~\cite{streampetr}; LiDAR: PointPillars~\cite{pointpillars}, CenterPoint~\cite{centerpoint}, VoxelNeXt~\cite{voxelnext}, TransFusion-L~\cite{transfusion}), raising the question of \textit{where the two still diverge}.

Fusion is not uniformly beneficial, and recent designs act on that. ECFusion~\cite{ecfusion} categorizes cross-modal conflicts into extrinsic and inherent types, and FDSNet~\cite{fdsnet} scores feature disagreement to select a fusion stage. Cocoon~\cite{cocoon} adjusts modality contributions via object-level uncertainty, and CCF~\cite{ccf} treats modality imbalance as a domain-generalization bottleneck. Sadeghian et al.~\cite{relifusion} derive per-modality reliability inside the detector and weight the fusion by it. All of these depend on knowing when to trust which modality. Yet each estimates that reliability inside a particular architecture, so the estimate is only as general as the network producing it. We derive ours outside any single detector, where it can be inspected before use.

\subsection{Fine-Grained Analysis of Failure}

Aggregate mAP says little about \emph{where} a detector fails on nuScenes~\cite{nuscenes}, KITTI~\cite{kitti}, or Waymo~\cite{waymo}. A body of work, therefore, breaks failure down along interpretable axes. TIDE~\cite{tide} splits detection error into six named types and scores each one's contribution to mAP. Graph-DETR3D~\cite{graphdetr3d} partitions 3D objects by camera viewpoint and finds localization error concentrated in overlapping regions. Safety-oriented evaluations reweight misdetection by its consequences for the ego vehicle~\cite{ceccarelli2022}. Corruption benchmarks stratify by weather, sensor noise, and calibration error~\cite{dong2023corruption,robustbenchfusion,multicorrupt}. Uncertainty studies trace confidence to distance and occlusion~\cite{feng2018uncertainty,feng2021review}. Rule mining links weather and lighting to crash types~\cite{ding2025}.

Each condition fails on an exogenous variable and then reports how performance varies along it. We follow that logic on a different quantity: the \emph{disagreement between two detectors}, measured per object and explained by its attributes. The unit is the single detection outcome, the granularity at which an adaptive fusion layer works.

\subsection{Attribution under Feature Dependence}
\label{sec:shaprw}

SHAP~\cite{shap} marginalizes features over a \textit{global} reference, assuming independence. Conditional sampling corrects the resulting bias when features are correlated~\cite{aas2021}. Such methods act \textit{globally}, producing one attribution profile per feature. Closest to our setting, Amoukou and Brunel~\cite{amoukou2026} make attribution regional by partitioning the input space. The partition there is learned. Driving supplies the partition already. The range is fixed by the sensors, and it is the axis along which their physics diverges. LiDAR returns thin with distance, while image resolution degrades on a different curve. Conditioning attribution on that axis yields an \textit{attribution transition matrix}, which names the dominant failure factor separately in each range band. SSFA constrains only the reference population, so it is composed of the estimators above (Sec.~\ref{sec:ssfa}).

\section{Methodology}

\subsection{Classification and Attribute Extraction}

Let $\mathcal{D} = \{(I_i, P_i, \mathcal{B}_i)\}_{i=1}^{N}$ be $N$ synchronized frames of camera images $I_i$, LiDAR point clouds $P_i$, and 3D box annotations $\mathcal{B}_i = \{b_j\}$.

\textbf{Perception state classification.} We deploy a camera-only detector $f_C$ (StreamPETR~\cite{streampetr}) and a LiDAR-only detector $f_L$ (TransFusion-L~\cite{transfusion}) to independently process each frame. For each ground-truth object $b_j$, we check whether either detector matches it, using BEV Intersection over Union (IoU) with $\text{IoU}_{\text{BEV}} > \tau$. That yields one of four states: BOTH, CAM\_ONLY, LIDAR\_ONLY, NEITHER. The label therefore depends on the detector pair; Sec.~\ref{sec:ablation} examines how well the structure survives a change in detector.

\textbf{Attribute extraction.} For each object $b_j$ we extract an attribute vector $\mathbf{a}_j$ in three groups. \textit{Spatial}: distance $d_j=\sqrt{x_j^2+y_j^2}$ and azimuth $\theta_j=\arctan(y_j/x_j)$. \textit{Sensor}: LiDAR point count $n_j^{\text{pts}}$ in the 3D box, annotated in nuScenes/Waymo and computed for KITTI, plus an observability level (nuScenes visibility 1--4; KITTI occlusion 0--3; Waymo difficulty). \textit{Object}: category $c_j$ and box volume $\text{vol}_j=l_j w_j h_j$. nuScenes carries the richest per-object metadata and is our primary analysis platform.

All records form the mining database $\mathcal{R} = \{(\text{CMPG}_j, \mathbf{a}_j)\}$, used for mining and training.

\textbf{CMPG-Bench.} The labeled database spans nuScenes (187{,}544 objects), Waymo ($\sim$191K) and KITTI ($\sim$19K), roughly 0.4M in all. It also covers the nuScenes train split used to fit the reliability field (Sec.~\ref{sec:mrm}). Because the label depends on the detector pair, we release CMPG-Bench as a reproducible labeling framework whose numbers move with that pair. It supports the six detectors used here and any others.

\subsection{Spatial Patterns and Association Rules}
\label{sec:patterndiscovery}

We mine $\mathcal{R}$ from two angles: \textit{where} the gaps concentrate and \textit{what attribute combinations} drive them.

\textbf{Spatial distribution analysis.} We partition the BEV plane into polar grids ($\ge$30 samples per cell). Let $N_{\texttt{X}}(g)$ count objects in state $\texttt{X}$ within cell $g$ and $N_{\text{total}}(g)$ their sum. Per-modality recall is then:
\begin{equation}
    R_C(g) = \frac{N_{\texttt{BOTH}}(g) + N_{\texttt{CAM\_ONLY}}(g)}{N_{\text{total}}(g)}
\end{equation}
\begin{equation}
    R_L(g) = \frac{N_{\texttt{BOTH}}(g) + N_{\texttt{LIDAR\_ONLY}}(g)}{N_{\text{total}}(g)}
\end{equation}
From these, we define two complementary spatial metrics. The \textit{Modality Dominance Score} (MDS) captures relative preference. The shared BOTH term cancels, leaving only the two exclusive states:
\begin{equation}
    \text{MDS}(g) = R_C(g) - R_L(g) \in [-1, 1]
\end{equation}
where positive values indicate higher camera and negative higher LiDAR reliability. The \textit{Joint Coverage} captures absolute detection capability:
\begin{equation}
    \text{Cov}(g) = 1 - \frac{N_{\texttt{NEITHER}}(g)}{N_{\text{total}}(g)} \in [0, 1]
\end{equation}
The two are complementary: MDS~$\approx$~0 cannot distinguish regions where both modalities succeed from regions where both fail, an ambiguity that Joint Coverage resolves.

\textbf{Association rule mining.} To find the attribute combinations behind the gaps, we discretize continuous attributes into semantic bins and mine rules with FP-Growth~\cite{fpgrowth}. We bin distance at 10, 30 and 50\,m and point count at 5, 20 and 50, four levels each. Visibility and category keep their labels. Minimum support is $\sigma_{\min} = 5/|\mathcal{R}|$, minimum confidence 0.5.

We keep rules with CMPG types as consequents, such as $\{\text{far}, \text{sparse}, \text{pedestrian}\} \Rightarrow \{\texttt{NEITHER}\}$ at support 0.031, confidence 0.82, and lift 2.63. Lift divides confidence by the consequent's base rate, so a value above one beats chance. A continuous account of nonlinear interactions complements this discrete summary, so we next fit a predictive model.

\subsection{Predictive Modeling and Interpretation}

\textbf{CMPG predictability.} We now move from discrete rules to a continuous model, training a LightGBM~\cite{lightgbm} four-class classifier on $\mathcal{R}$ with early stopping and balanced class weights. Folds are stratified by scene rather than by object, which prevents leakage from intra-scene temporal correlations. The classifier confirms that the state is predictable and is the model we interpret below.

\textbf{SSFA.} Standard SHAP~\cite{shap} marginalizes over a global reference distribution, which in spatially non-stationary data yields misleading attributions. Evaluating point-count importance for a far pedestrian requires reference values from nearby trucks with hundreds of points, which is impossible at 45\,m.

SSFA instead conditions the reference distribution on the spatial zone. It shares the conditional-reference idea explored for dependent features~\cite{aas2021}, but applies it as a diagnostic (Sec.~\ref{sec:shaprw}). Let $f$ denote the trained LightGBM model, $\mathcal{F} = \{1, \ldots, K\}$ the $K$ attributes, and $\{r_1, \ldots, r_M\}$ the distance strata ($M\!=\!4$: 0--10, 10--30, 30--50, 50+\,m). For an object $j$ in stratum $r$, SSFA computes the Shapley value of attribute $k$ using only co-located objects as reference:
\begin{equation}
    \phi_k^{(r)}(j) = \!\sum_{S \subseteq \mathcal{F} \setminus \{k\}} w_S \left[\mathbb{E}_r[f | \mathbf{a}_{S \cup k}] - \mathbb{E}_r[f | \mathbf{a}_S]\right]
\end{equation}
where $w_S = |S|!(K\!-\!|S|\!-\!1)!/K!$ are the standard Shapley coalition weights and $\mathbf{a}_S$ is the attribute subset indexed by $S$. The expectation $\mathbb{E}_r[\cdot]$ is taken over the spatially restricted reference set $\mathcal{R}^{(r)} = \{i \mid d_i \in \text{range } r\}$, on the predicted probability of the object's CMPG class. SSFA restricts only the reference data to zone $r$, leaving $f$ itself global. The per-stratum mean absolute attributions form the \textit{attribution transition matrix}:
\begin{equation}
    \Phi_{rk} = \frac{1}{|\mathcal{R}^{(r)}|}\sum_{j \in \mathcal{R}^{(r)}} \left|\phi_k^{(r)}(j)\right|
\end{equation}
where $\Phi \in \mathbb{R}^{M \times K}$ indexes strata by row and attributes by column, showing how the dominant factors shift across zones.

\subsection{Modality Reliability Map for Fusion}
\label{sec:mrm}

SSFA shows that the factors governing disagreement shift systematically across the BEV plane, which dictates how a reliability prior should be constructed. A single global regressor would average the zones together and blur exactly what SSFA exposes. The Modality Reliability Map (MRM) therefore fits one regressor per distance stratum. A lightweight layer injects the resulting field into existing feature-level fusion detectors.

\textbf{From per-object reliability to a spatial field.} The per-object attributes are available only \emph{after} detection, whereas feature-level fusion operates on BEV maps \emph{before} objects exist. We therefore reduce the estimate to quantities available at every cell beforehand: distance $d=\sqrt{x^2+y^2}$, azimuth $\theta=\arctan(y/x)$, and online point density $\rho(x,y)$. Zone-specific regressors $\{g^{(r)}\}$ map these detection-free inputs to the per-cell recalls $R_C, R_L$ of Sec.~\ref{sec:patterndiscovery}. Each cell thus receives a camera and a LiDAR reliability score,
\begin{equation}
    S_C(x,y) = g^{(r)}_C(d,\theta,\rho), \quad
    S_L(x,y) = g^{(r)}_L(d,\theta,\rho),
\end{equation}
where $r$ is the zone containing $(x,y)$. The pair $\{S_C, S_L\}$ forms a two-channel reliability field at the fusion BEV resolution, regressed independently of the per-object CMPG classifier. All three inputs are computable online from the sensor stream. Ground-truth attributes are used only to generate training labels for the regressors, and the deployed model never observes them. Being an explicit function of three interpretable quantities, the field can also be audited before deployment. It can be rendered as a map, queried at any location to determine the reliability it would assert there, and checked against the rules in Table~\ref{tab:decision}. A learned fusion weight admits none of these checks.

\textbf{Split discipline.} We fit the regressors $\{g^{(r)}\}$ only on CMPG labels from the nuScenes \emph{train} split, disjoint from the val split where the fusion hosts are evaluated (Sec.~\ref{sec:mrmresults}). The analyses in Sec.~\ref{sec:results} characterize the val split and never build the field, so no validation statistics enter the prior.

\textbf{Prior-Guided Spatial Attention (PGSA).} Most BEV-space fusion detectors expose two modality-specific BEV feature maps $F_C(x,y), F_L(x,y)$ before merging them, on which PGSA operates directly (e.g., BEVFusion, GAFusion). Some hosts keep their camera branch outside BEV, among them TransFusion and FUTR3D, where object queries consume image features through cross-attention. There, we first lift the camera features onto the shared BEV grid with the host's own projection, the standard lift-splat operation~\cite{bevfusion}, and then apply PGSA. In either case, PGSA reweights the two streams by their per-cell reliability and hands them to the host's fusion operator unchanged. Where self-attention scales quadratically with the number of BEV cells, PGSA runs a per-cell two-way competition and scales linearly. For each cell, it forms one score per modality, combining a light feature projection with the reliability prior,
\begin{equation}
    s_m(x,y) = \mathbf{w}_m^\top F_m(x,y) + \lambda \log S_m(x,y), \quad m\in\{C,L\},
    \label{eq:pgsa_score}
\end{equation}
turns the scores into per-cell modality weights, and rescales each stream,
\begin{equation}
\begin{aligned}
    \alpha_m(x,y) &= \frac{e^{s_m(x,y)}}{e^{s_C(x,y)}+e^{s_L(x,y)}}, \\[2pt]
    \tilde F_m(x,y) &= \alpha_m(x,y)\, F_m(x,y),
\end{aligned}
\end{equation}
after which $\tilde F_C, \tilde F_L$ enter the host's native fusion. Here, $\mathbf{w}_C,\mathbf{w}_L$ are learned projection vectors and $\lambda$ is a learned scalar controlling prior strength. We floor $S_m$ at $10^{-3}$ so that the logarithm stays finite where a modality's predicted recall approaches zero. Entering the logits additively, the prior multiplies the feature-driven weight by $S_m^{\lambda}$ after the softmax, so $\lambda$ sets how sharply reliability rescales the gate. At $S_C\!=\!S_L$, the two factors are equal, and the layer reduces to a feature-driven gate.

\textbf{Plug-in training.} We precompute the field once from train-split labels of the primary StreamPETR$\times$TransFusion-L pair and reuse it across hosts. We insert PGSA into four pretrained detectors spanning distinct paradigms (FUTR3D~\cite{futr3d}, BEVFusion~\cite{bevfusion}, TransFusion~\cite{transfusion}, GAFusion~\cite{gafusion}). Backbones stay frozen while we fine-tune the fusion stage, learning $\lambda$ and $\mathbf{w}_C,\mathbf{w}_L$ per host. Two controls under the same schedule separate what the prior contributes from what the fine-tuning does. The \emph{no-prior} variant drops the $\lambda\log S_m$ term, leaving only the feature-driven gate. The \emph{reliability-weighting} variant applies $S_m$ as a per-cell weight without $\mathbf{w}_m$ or $\lambda$. Fig.~\ref{fig:pipeline} shows the pipeline.

\subsection{Statistical Validation and Robustness}

We assess discrete attributes with Bonferroni-corrected Chi-squared tests and continuous ones with Kruskal-Wallis H tests. With $|\mathcal{R}|>10^5$ the $p$-value alone is uninformative, so we report Cram\'{e}r's V and $\eta^2$ as effect sizes. Robustness covers three parameter axes ($\tau$, confidence threshold, binning scheme), two distance partitions, a detector sweep, and transfer to KITTI~\cite{kitti} and Waymo~\cite{waymo}.

\section{Experimental Setup}

\subsection{Datasets}

\textbf{nuScenes}~\cite{nuscenes} is the primary platform (1,000 scenes, 6$\times$camera + 32-beam LiDAR, 6,019 val frames). \textbf{KITTI}~\cite{kitti} adds a denser 64-beam LiDAR but only a front view. \textbf{Waymo}~\cite{waymo} uses a different rig (5$\times$LiDAR + 5$\times$camera); we sample $\sim$4{,}000 keyframes ($\sim$191K objects).

\subsection{Detectors}

All detectors use officially released pretrained weights. The primary pair is StreamPETR (camera, 55.0\% mAP) and TransFusion-L (LiDAR, 65.5\%). The ablation in Sec.~\ref{sec:ablation} sweeps the camera detector across 29.8--55.0\% mAP, using FCOS3D, BEVFormer and StreamPETR. For the fusion hosts we report val mAP under our own protocol, without test-time augmentation (TTA) or ensemble. Numbers may therefore differ slightly from the original papers. They are FUTR3D 67.3\%, BEVFusion 68.5\%, TransFusion 67.5\% and GAFusion 72.1\%.

\subsection{Configuration and Metrics}

Both the BEV IoU and confidence thresholds default to 0.3. The sensitivity analysis sweeps $\tau \in \{0.25, 0.5\}$ and confidence $\in \{0.2, 0.4\}$. FP-Growth uses $\sigma_{\min} = 5/|\mathcal{R}|$ and minimum confidence 0.5. LightGBM runs 5-fold scene-stratified CV with early stopping (patience 50), balanced class weights, 255 leaves, learning rate 0.05 and up to 1{,}000 rounds. The BEV grid is $\sim$2$\times$2\,m, and all runs use dual RTX Pro 4000 GPUs.

Analysis metrics are 5-fold scene-stratified CV means with standard deviations. The fusion-injection results (Sec.~\ref{sec:mrmresults}) use five random-seed runs.

\section{Results and Analysis}
\label{sec:results}

\subsection{CMPG Distribution}

Table~\ref{tab:cmpg_dist} presents the CMPG distribution on nuScenes val (187{,}544 ground-truth objects). All attribute--CMPG associations are significant (Chi-squared $p<0.001$, Cram\'{e}r's V $>0.15$; Kruskal-Wallis $p<0.001$, $\eta^2>0.1$). CAM\_ONLY rates are inversely correlated with physical cross-section, ordering cones (16.8\%) $>$ bicycles (14.2\%) $>$ pedestrians (13.7\%) $\gg$ cars (5.3\%) $>$ trucks (2.9\%). The rate peaks at 10--30\,m (11.3\%) instead of falling monotonically. In that band the LiDAR point cloud has already grown sparse, while image resolution is still sufficient for the camera to detect small objects. The highest camera-only rates belong to the small categories, and that regime survives inside the range assumed to be LiDAR's.

\begin{table}[!t]
\caption{CMPG distribution (\%) on nuScenes val (StreamPETR $\times$ TransFusion-L). Bold marks the dominant state in each distance band.}
\label{tab:cmpg_dist}
\centering
\footnotesize
\setlength{\tabcolsep}{4pt}
\begin{tabular}{lcccc}
\toprule
& \textbf{BOTH} & \textbf{CAM\_O} & \textbf{LID\_O} & \textbf{NEI} \\
\midrule
Overall & 37.8 & 8.6 & 22.4 & 31.2 \\
\midrule
Car & 52.1 & 5.3 & 19.6 & 23.0 \\
Pedestrian & 28.4 & 13.7 & 20.8 & 37.1 \\
Truck & 46.5 & 2.9 & 24.2 & 26.4 \\
Bicycle & 21.3 & 14.2 & 18.5 & 46.0 \\
Traffic cone & 30.2 & 16.8 & 14.3 & 38.7 \\
\midrule
0--10\,m & \textbf{70.9} & 10.4 & 13.5 & 5.2 \\
10--30\,m & \textbf{46.2} & 11.3 & 22.8 & 19.7 \\
30--50\,m & 30.2 & 7.7 & 26.9 & \textbf{35.2} \\
50+\,m & 19.1 & 3.8 & 18.8 & \textbf{58.3} \\
\bottomrule
\end{tabular}
\end{table}

\begin{figure*}[t]
\centering
\includegraphics[width=0.72\textwidth]{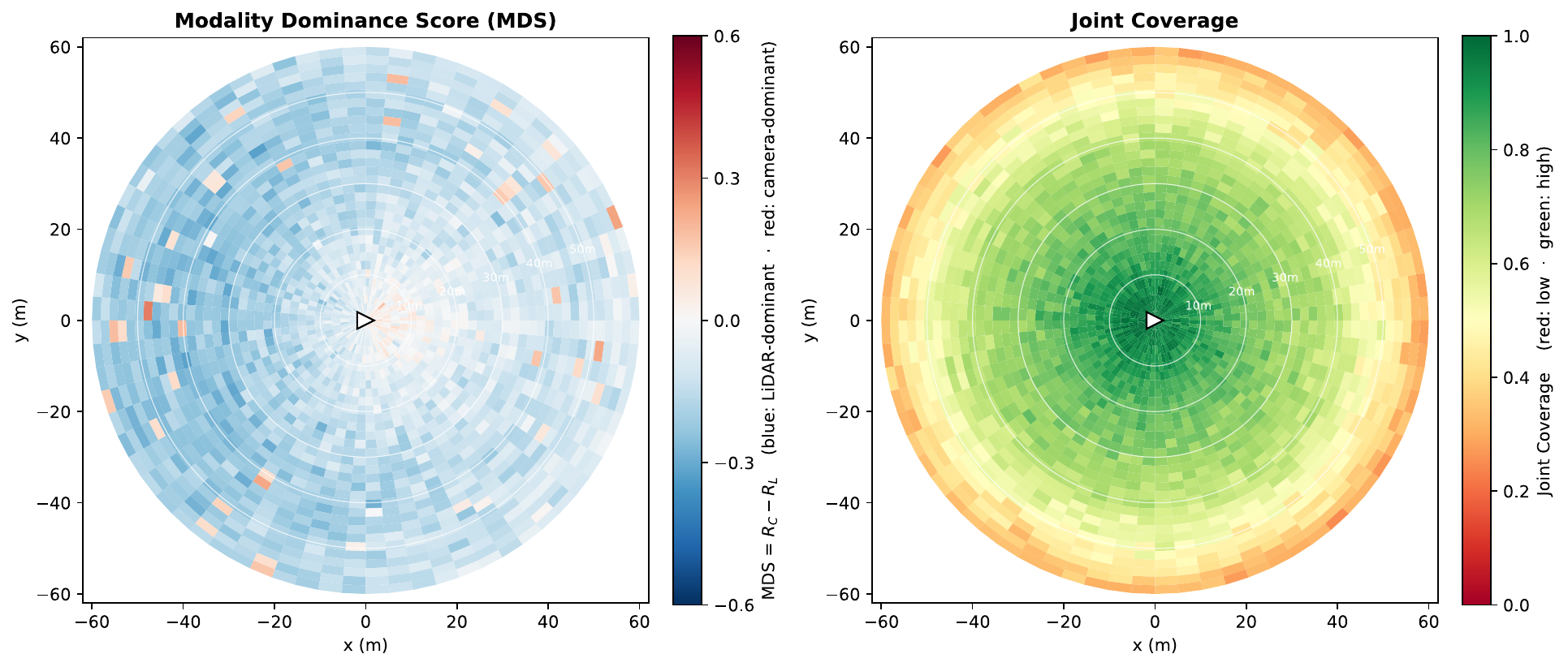}
\caption{BEV spatial analysis (StreamPETR $\times$ TransFusion-L, nuScenes val). \textbf{Left:} MDS heatmap, blue LiDAR-dominant, red camera-dominant. \textbf{Right:} Joint Coverage, green high, red low. The front-to-rear asymmetry is weak on this colour scale and is quantified per sector in Sec.~\ref{sec:spatial}.}
\label{fig:mds_cov}
\end{figure*}

\subsection{MDS and Joint Coverage}
\label{sec:spatial}

Fig.~\ref{fig:mds_cov} presents the two-channel BEV heatmap. The core is near-parity, with mean MDS $\approx-0.03$ within 10\,m. LiDAR dominance then \emph{strengthens} with distance, reaching $-0.12$ at 10--30\,m and peaking at $-0.19$ at 30--50\,m. It eases to $-0.15$ beyond 50\,m as both degrade. Joint Coverage falls monotonically from $\sim$0.95 within 10\,m to $\sim$0.42 beyond 50\,m. Adaptive fusion gains most in the mid-range annulus, where disagreement peaks while coverage stays substantial.

\textbf{Directional non-stationarity.} The map is not radially symmetric. Mean MDS falls from $-0.07$ frontally ($\pm45^\circ$) to $-0.20$ in the rear ($135$--$225^\circ$). That 0.13 swing rivals an entire distance stratum. A likely cause is sensor geometry, the nuScenes rear camera using a wider field-of-view lens. Being orthogonal to distance, the asymmetry is invisible to the distance-only analysis common in fusion work.

\subsection{From Rules to Decisions}

FP-Growth extracts 247 rules with a CMPG state as the consequent. The highest-lift ones are diagnostic. $\{$near, very sparse, cone$\}\Rightarrow$CAM\_ONLY reaches lift 7.79. Far-range trucks resolve to LIDAR\_ONLY instead of NEITHER (lift 2.59), their large surface maintaining returns. Both are joint conditions, not single-variable ones.

These rules collapse into the decision table of Table~\ref{tab:decision}, which replaces single-variable maxims with \textit{joint} thresholds on point count, distance and category. The table is a model-free fallback. The Modality Reliability Map (Sec.~\ref{sec:mrm}) generalizes it continuously.

\begin{table}[tb]
\caption{Decision rules distilled from CMPG mining. Confidences are read off the supporting association rules. pts: LiDAR point count; $d$: distance.}
\label{tab:decision}
\centering
\footnotesize
\setlength{\tabcolsep}{3pt}
\begin{tabular}{@{}p{4.3cm}p{2.0cm}c@{}}
\toprule
\textbf{Condition} & \textbf{Trust} & \textbf{Conf.} \\
\midrule
pts $<5$, class $\in$\{cone, ped\} & camera (CAM\_ONLY) & .54--.67 \\
$d>50$\,m \textbf{and} pts $<5$ & low-confidence & .91 \\
$d>50$\,m, pts $\in[5,20]$, large class & LiDAR (LID\_ONLY) & .58 \\
$d<30$\,m, pts $\ge50$, class $=$ car & both (BOTH) & .89 \\
\bottomrule
\end{tabular}
\end{table}

\begin{figure}[tb]
\centering
\includegraphics[width=0.95\columnwidth]{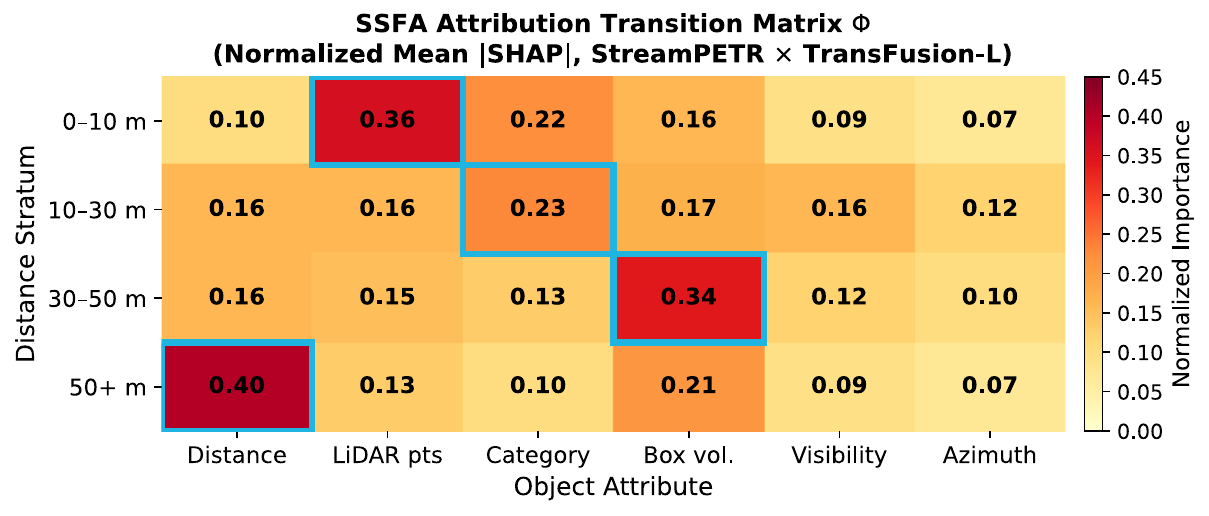}
\caption{SSFA attribution transition matrix $\Phi$. Rows are distance strata, columns are object attributes, and color is the row-normalized mean $|\phi|$. Boxed cells mark the leading factor in each stratum.}
\label{fig:ssfa}
\end{figure}

\subsection{SSFA Analysis}
\label{sec:ssfa}

\textbf{Attribution transition matrix.} Fig.~\ref{fig:ssfa} presents $\Phi$, each row normalized so importances sum to one within a zone. LiDAR point count dominates at 0--10\,m (0.36), category rises at 10--30\,m (0.23), box volume overtakes at 30--50\,m (0.34), and distance itself dominates beyond 50\,m (0.40). The factor driving failure differs in each zone.

\textbf{Adding the angular axis.} Modality dominance varies with viewing direction (Sec.~\ref{sec:spatial}), so we re-run SSFA on a two-dimensional partition. It crosses the four distance strata with a frontal ($\pm45^\circ$) and a rear ($135$--$225^\circ$) sector. Two things follow. The azimuth column collapses from $0.07$--$0.12$ to $0.03$--$0.07$, since direction is now a stratification variable with no within-stratum variation left to explain. The near$\to$far top-1 sequence, meanwhile, survives in both sectors unchanged from Fig.~\ref{fig:ssfa}. What shifts is the weight behind that ordering. The rear sector has MDS $-0.20$ against $-0.07$ frontally. Their point count carries $0.19$ of the 10--30\,m row against $0.17$ at the front, and distance $0.18$ against $0.16$. Direction modulates how strongly each factor acts, not which one dominates. The distance partition captures the primary structure; the angular axis refines it.

\textbf{Comparison with global SHAP.} Pooled over all objects, global attribution ranks point count $>$ distance $>$ category, an ordering that holds nowhere in particular. SSFA reveals a zone-dependent reordering, with distance only 4th within 0--10\,m. Global attribution conflates the strong between-zone effect of distance with its weak within-zone effect.

\textbf{Faithfulness.} We mask each method's top-3 features with zone-mean values and measure the prediction shift. SSFA yields the largest shift in all four strata, with $\Delta=0.193/0.179/0.197/0.201$ from near to far and std $\le0.016$. That exceeds conditional-global SHAP~\cite{aas2021} by 5--34\%, with paired $t_4$ from 3 to 10 and all $p<0.05$. The subscript denotes four degrees of freedom from five folds. Swapping the inner estimator for the conditional sampler of~\cite{aas2021} preserves the per-zone ranking (Spearman $\rho=0.96/0.90/0.91/0.94$), so the reordering is a property of the data. This protocol is not neutral. Zone-mean masking is itself zone-conditional, so it measures fidelity to zone-local behaviour rather than general attribution quality.

\textbf{Non-additivity.} Under the distance partition, the distance$\times$point-count interaction (mean $|\phi_{ij}|=0.087$) is the strongest off-diagonal term and exceeds azimuth's entire main effect (0.049). The joint far-and-sparse condition therefore degrades detectability beyond the sum of its parts. The point-count$\times$category interaction (0.058) localizes the camera-only regime to small classes. Category$\times$visibility (0.031) fits uncertainty driven by distance and occlusion~\cite{feng2018uncertainty}.

\subsection{Predictive Performance}

The global LightGBM achieves macro-F1 = 0.691 $\pm$ 0.007, with per-class scores of 0.76 for BOTH, 0.45 for CAM\_ONLY, 0.74 for LIDAR\_ONLY, and 0.81 for NEITHER. Zone-specific models do best at the extreme zones, reaching 0.913 at 0--10\,m and 0.894 beyond 50\,m. They do worst in the mid-range, 0.671 and 0.642, where interactions are strongest. CAM\_ONLY is the hardest state to predict, which matters because the camera-trust rules rest on it. At 8.6\% it is the rarest state and the least separable, arising at the boundary of LiDAR detectability for small objects. Those rules carry low confidence, .54--.67 against .89 for the BOTH row, and work best as a soft preference, not a hard switch. The continuous field needs no commitment to a discrete state and is the safer basis for deployment.

\subsection{MRM Fusion Validation}
\label{sec:mrmresults}

We evaluate the MRM prior as a plug-in on four pretrained fusion detectors. We freeze the backbones and fine-tune only the fusion stage on the nuScenes train split, then evaluate on val without TTA. The field is fitted on train-split labels only (Sec.~\ref{sec:mrm}), so neither the prior nor the fine-tuning sees val data. PGSA itself adds only $\mathbf{w}_C,\mathbf{w}_L,\lambda$. At inference, PGSA adds 0.5\,ms per frame on an RTX Pro 4000, an increase of 0.4\% over the unmodified host. Baselines are the official pretrained checkpoints under the same protocol. We report means over five seeds and assess gains with paired $t$-tests.

\begin{table}[tb]
\caption{MRM injected through PGSA into each pretrained host (nuScenes val, fine-tuned on train, no TTA; means over five seeds, per-cell std $0.12$--$0.31$). The two BEVFusion controls are defined in Sec.~\ref{sec:mrm}.}
\label{tab:mrm}
\centering
\footnotesize
\setlength{\tabcolsep}{3pt}
\begin{tabular}{@{}lcccc@{}}
\toprule
\textbf{Detector} & \textbf{mAP} & \textbf{NDS} & \textbf{$\Delta$mAP} & \textbf{$\Delta$NDS} \\
\midrule
Late fusion (score-level MRM) & 66.7 & 69.9 & $-$ & $-$ \\
\midrule
FUTR3D~\cite{futr3d} & 67.3 & 68.0 & $-$ & $-$ \\
\quad + MRM (ours) & \textbf{67.7} & \textbf{68.3} & +0.4 & +0.3 \\
BEVFusion~\cite{bevfusion} & 68.5 & 71.4 & $-$ & $-$ \\
\quad + fine-tune, no prior & 68.8 & 71.6 & +0.3 & +0.2 \\
\quad + reliability weighting & 69.2 & 71.8 & +0.7 & +0.4 \\
\quad + MRM (ours) & \textbf{69.6} & \textbf{72.0} & +1.1 & +0.6 \\
TransFusion~\cite{transfusion} & 67.5 & 71.3 & $-$ & $-$ \\
\quad + MRM (ours) & \textbf{68.8} & \textbf{72.1} & +1.3 & +0.8 \\
GAFusion~\cite{gafusion} & 72.1 & 73.5 & $-$ & $-$ \\
\quad + MRM (ours) & \textbf{72.2} & 73.5 & +0.1 & +0.0 \\
\bottomrule
\end{tabular}
\end{table}

Table~\ref{tab:mrm} reports the main result. The prior improves BEVFusion by +1.1 mAP and TransFusion by +1.3 mAP with backbones frozen, consistently across seeds ($t_4=9.5$ and $14.2$, $p<0.01$). For FUTR3D, whose query-based sampler already performs some adaptive fusion, the gain is smaller but significant at +0.4 mAP ($t_4=5.5$, $p<0.05$). For GAFusion, it is not significant, +0.1 mAP with $t_4=1.9$. The nuScenes Detection Score (NDS) gains less, +0.0 to +0.8. The prior adjusts per-cell weighting, not the regression terms NDS also scores. Score-level late fusion reaches 66.7 mAP.

The gain decreases as the host's native fusion becomes more adaptive, with the order TransFusion $>$ BEVFusion $>$ FUTR3D $>$ GAFusion. A host that already learns reliability implicitly leaves less headroom for an explicit prior. Two consequences follow. First, the pattern indicates that the prior carries a reliability structure and does not act as generic regularization. For BEVFusion, the scope of a random field of matched magnitude does not help at all. The gain across hosts instead tracks how much structure each host is missing. Second, the ordering delimits where the prior is useful. The gain is largest for hosts with fixed or lightly adaptive fusion, which are still the most deployed detectors. On a guided-adaptive host, the prior instead offers an auditable version of what that host learned implicitly.

\textbf{Isolating the prior.} The two controls of Sec.~\ref{sec:mrm} separate the prior's own contribution from that of the fine-tuning. The no-prior variant recovers only +0.3 mAP, placing the bulk of the +1.1 with the injected structure and not with the extra optimization. Reliability weighting recovers +0.7 mAP, so the mined structure is useful on its own. The attention formulation adds +0.4 more by letting the host modulate how far the prior is trusted.

\textbf{What in the prior matters.} Table~\ref{tab:mrm_ablation} isolates each ingredient on BEVFusion. Online point density alone accounts for +0.6 of the +1.1, acting as a detection-free proxy for sparsity. A random field or an overly strong fixed $\lambda$ does not help. The random field even falls below the no-prior control, so an uninformative prior interferes with the host's own fusion. What helps is the structure, not the signal path.

\begin{table}[htbp]
\caption{Ablation of the reliability field on BEVFusion ($\Delta$mAP over the 68.5 baseline). Rand: random field; Pos: position only ($d,\theta$); +$\rho$: adding online point density; Full: full zone-specific field (learned $\lambda$); Over: over-strong fixed $\lambda$.}
\label{tab:mrm_ablation}
\centering
\footnotesize
\setlength{\tabcolsep}{5pt}
\begin{tabular}{@{}lccccc@{}}
\toprule
& \textbf{Rand} & \textbf{Pos} & \textbf{+$\rho$} & \textbf{Full} & \textbf{Over} \\
\midrule
$\Delta$mAP & $-0.2$ & $+0.2$ & $+0.6$ & $\mathbf{+1.1}$ & $+0.5$ \\
\bottomrule
\end{tabular}
\end{table}

\subsection{Ablation Studies}
\label{sec:ablation}

\textbf{Detector ablation.} Sweeping each modality in turn separates sensor-level regularities from detector-specific artifacts. Varying the camera detector holds $R_L$ at 60.2, and varying the LiDAR detector holds $R_C$ at 46.4. As either modality strengthens, BOTH rise, and NEITHER falls. The relative structure is invariant under both sweeps. Pedestrian CAM\_ONLY exceeds car CAM\_ONLY at every camera level, 6.4 against 2.6 for FCOS3D, 9.5 against 3.7 for BEVFormer, and 13.7 against 5.3 for StreamPETR. The SSFA top-1 stays point count at 0--10\,m and distance at 50+\,m for every pair.

\textbf{Parameter sensitivity.} The findings are stable under the protocol choices. We vary the IoU threshold to $0.25$ and $0.5$, the confidence threshold to $0.2$ and $0.4$, and the binning to equal-frequency. The top rule set retains Jaccard $\ge0.71$ with the default, and macro-F1 stays within $0.69\pm0.01$. The strata boundaries follow LiDAR density transitions: dense returns within $\sim$10\,m, thinning past $\sim$30\,m, and the joint range limit near $\sim$50\,m. Re-running SSFA under the partitions $\{5,20,40\}$ and $\{15,35,55\}$ leaves the near$\to$far top-1 sequence unchanged. The reordering is not an artifact of the boundaries.

\subsection{Cross-Dataset Validation}

The sensor suites differ substantially, from 32 to 64 LiDAR beams and from surround to front-only cameras. The structure that Table~\ref{tab:cross_dataset} can test nonetheless holds on every dataset. Pedestrians show higher CAM\_ONLY rates than cars, supporting the category dependence of Insight~1. The SSFA top-1 factor matches within each stratum, so the zone-wise reordering is not a nuScenes artifact. Insights~2 and~3 need the full attribute vector and per-distance breakdown, so they are evaluated on nuScenes only. Retrained F1 is highest on KITTI at 0.718, whose denser 64-beam LiDAR gives cleaner point-count signals, and lowest on Waymo at 0.672. Zero-shot transfer reaches 0.648/0.612 on KITTI/Waymo, 90--91\% of retrained, so the patterns reflect sensing geometry rather than dataset artifacts.

\begin{table}[htbp]
\caption{Cross-dataset validation. Zero-shot: the nuScenes-trained LightGBM applied without retraining.}
\label{tab:cross_dataset}
\centering
\footnotesize
\setlength{\tabcolsep}{4pt}
\begin{tabular}{lccc}
\toprule
& \textbf{nuScenes} & \textbf{KITTI} & \textbf{Waymo} \\
\midrule
Retrained F1 & 0.691 & 0.718 & 0.672 \\
Zero-shot F1 & $-$ & 0.648 & 0.612 \\
\midrule
Ped CAM\_ONLY & 13.7\% & 11.2\% & 14.8\% \\
Car CAM\_ONLY & 5.3\% & 3.8\% & 5.1\% \\
\midrule
SSFA top-1 (0--10\,m) & pts & pts & pts \\
SSFA top-1 (50+\,m) & distance & distance & distance \\
\bottomrule
\end{tabular}
\end{table}

\FloatBarrier
\section{Discussion}

\textbf{Towards reliability-aware 3D perception.} A safety case rests on arguments, and a fusion detector that cannot say where it trusts which sensor offers nothing to argue from. CMPG supplies the missing vocabulary. It records which modality succeeded for every object. Those records aggregate into a field stating, before detection, how far each sensor can be trusted at each location. Where mAP is one number for a whole suite, the field is a claim. It is checkable cell by cell, refutable by a counterexample, and revisable when the sensors change. Certifying multi-modal perception demands a description at this level, and CMPG-Bench offers a starting point. The same reasoning points beyond detection. 3D occupancy prediction~\cite{surroundocc,occ3d} exposes free space and visibility online, and a field defined over occupancy would carry the argument into what planning actually consumes.

\textbf{Scaling the evidence.} The present database holds 0.4M objects from three sensor suites, labelled by one detector pair. Rare categories and long-tail conditions appear too thinly in it to support rules of their own. Broadening it means running the same pipeline across additional detector pairs, more datasets, and, eventually, radar and thermal channels. The field sharpens as the evidence behind each cell grows. The attribute vector widens the same way. Illumination and weather are absent by construction. They vary per scene, not per object. A scene-level tag cannot explain why the camera found one pedestrian in a rainy frame and missed the next. Reaching those modes needs a per-object exposure or reflectance estimate, which no dataset annotates. Our attributes, therefore, speak to geometry, and the environmental axis stays open. The partition can grow too. Sec.~\ref{sec:ssfa} already stratifies attribution by direction as well as range. Carrying that axis into the reliability field itself is the next step.

\textbf{Present scope.} Three boundaries delimit the current results. We read the labels through a detector pair representative of current practice, then ask what survives when that anchor moves. Absolute proportions shift with detector capability, but the relative structure holds across a 25-point mAP sweep (Sec.~\ref{sec:ablation}). A stronger pair rescales the numbers without invalidating the ordering. Among the paradigms of Sec.~\ref{sec:fusion3d}, fully sparse detectors expose no dense BEV grid and remain untested. A detector can miss an object, hallucinate one, or place it imprecisely. Our label isolates the first, asking only whether a detector matched a given ground-truth object. CMPG therefore measures reliability in the sense of recall. False positives and localization quality would each need their own state definition, which this construction supports. 

\textbf{Relation to uncertainty-based fusion.} Our reliability prior is an \emph{offline}, spatially mined signal, whereas methods such as Cocoon~\cite{cocoon} estimate \emph{online} per-object uncertainty~\cite{feng2021review}. The two are complementary. Because PGSA injects the prior as an additive term in its per-cell logits (Eq.~\ref{eq:pgsa_score}), an online uncertainty score can sit alongside it. Whether the combination pays off remains to be measured.

\section{Conclusion and Future Work}

This paper presented the CMPG framework for mining cross-modal perception gaps between camera and LiDAR detectors. Mining 0.4M labelled objects turns three qualitative assumptions of the field into quantitative statements. Modality preference follows object category rather than distance. Near range does not favor LiDAR for small objects. The drivers of failure interact rather than add. The category dependence is reproduced across nuScenes, KITTI, and Waymo, while the other two rely on the richly annotated nuScenes split. Modality dominance also varies with viewing direction, a second spatial axis alongside range. We distill these patterns into explicit decision rules. Injecting the resulting Modality Reliability Map through a prior-guided spatial-attention layer improves feature-level fusion detectors. Beyond these findings, CMPG-Bench reframes multi-modal benchmarks. Instead of comparing detectors solely by aggregate mAP, it reveals \emph{where} and \emph{why} modalities disagree for each object, so that the reliability structure, not architecture alone, can improve perception.
As part of future work, we will carry the angular axis into the reliability field itself and define that field over 3D occupancy, so that reliability reaches the representation planning consumes.


\end{document}